\documentclass[11pt]{article}

\usepackage[preprint]{acl}
\usepackage{times}
\usepackage{latexsym}
\usepackage[T1]{fontenc}
\usepackage[utf8]{inputenc}
\usepackage{microtype}
\usepackage{inconsolata}
\usepackage{graphicx}
\usepackage{booktabs}
\usepackage{amsmath,amssymb,amsthm,mathtools}
\usepackage{enumitem}
\usepackage{xcolor}
\usepackage{url}
\usepackage{multirow}

\newtheorem{proposition}{Proposition}

\newtheorem{corollary}{Corollary}
\theoremstyle{definition}
\newtheorem{definition}{Definition}

\newcommand{\score}{\phi}
\newcommand{\surp}{\mathcal{I}}
\newcommand{\model}{q_\theta}
\newcommand{\attackname}{\textsc{Specificity Padding}}
\newcommand{\task}{\mathcal{T}}
\newcommand{\histories}{\mathcal{H}}
\newcommand{\observations}{\mathcal{O}}
\newcommand{\ctx}{\textsc{Matched}}
\newcommand{\hid}{\textsc{Hidden}}
\newcommand{\wrg}{\textsc{Wrong}}
\newcommand{\StrictOrderN}{23}
\newcommand{\StrictOrderTotal}{30}

\newcommand{\BaselinePolish}{0.846}
\newcommand{\BaselineSummary}{0.708}
\newcommand{\BaselineTitle}{0.536}
\newcommand{\BaselineSubject}{0.488}

\newcommand{\CrossFixedMin}{0.324}
\newcommand{\CrossFixedMax}{0.505}

\title{Likelihood Is Not Authorship:\\
Can We Reliably Measure Human Contribution in AI-Assisted Writing?}

\newcommand{\arxivauthornames}{Hasan Amin, Zhuoran Lu, Ming Yin}
\newcommand{\arxivaffiliationone}{Department of Computer Science, Purdue University}
\newcommand{\arxivhasanemail}{hasanamin@purdue.edu}
\newcommand{\arxivzhuoranemail}{zhuoranlu33@gmail.com}
\newcommand{\arxivmingemail}{mingyin@purdue.edu}
\author{Hasan Amin \quad Zhuoran Lu \quad Ming Yin \\
  \normalfont\arxivaffiliationone \\
  \normalfont\texttt{\arxivhasanemail, \arxivzhuoranemail, \arxivmingemail}}

\providecommand{\arxivauthornames}{}
\hypersetup{
  pdftitle={Likelihood Is Not Authorship: Can We Reliably Measure Human Contribution in AI-Assisted Writing?},
  pdfauthor={\arxivauthornames},
  pdfkeywords={human contribution, AI-assisted writing, likelihood, provenance}
}

\begin{document}
\maketitle

\begin{abstract}
AI-text detection generally asks whether a machine wrote a text.
As AI becomes an increasingly integral and accepted part of writing, the more relevant question is no longer whether AI was used, but how much of the text reflects human contribution.
A prominent line of work estimates this by measuring how predictive a person's prompt is of the final text. 
We show that this measure can be easily manipulated.
We introduce \attackname{} that simply asks the model to copy a few random IDs from the prompt, and that alone makes a title-only request look more human than one that supplies the article's key ideas.
A context-swap audit then holds each document fixed, and only shows or hides the IDs to the evaluator. The score of the unchanged text swings by a third to half of its scale, and it rises for every document we test, on news and scientific abstracts, across varied LLMs from 3.8B to 753B parameters. The same articles can average 85\% human or 15\% human depending only on which prompt is on record, and even with no prompt at all, the output-only variant rewards the random strings themselves. An exact identity, the credit ledger, explains why. The score is a surprisal-weighted average over the text, so rare strings that the context makes copyable collect the credit, whatever they mean. A two-point bound marks a deeper limit. If the prompt could have been machine-written, every transcript-only method must, in the worst case, miss by half the contribution at stake. 
Likelihood-based scores alone cannot reliably establish human contribution without evidence of what the person added and how the text was produced.
\end{abstract}

\section{Introduction}

\begin{figure}[t]
    \centering
    \includegraphics[width=\columnwidth]{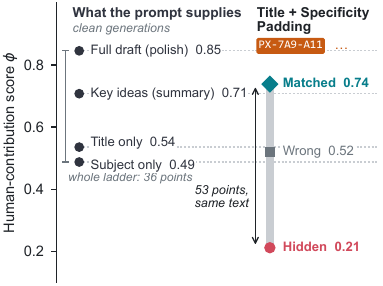}
    \vspace{-16pt}
    \caption{\textbf{Copying a few random IDs outranks supplying the key ideas.} Left: the likelihood-based human-contribution score \citep{xie2026human} falls as the prompt supplies less of the article. Right: articles generated from a title-only request with \attackname{}, which asks the model to copy eight random IDs. Scored with the IDs in the evaluator's context (\ctx{}), they rank above articles written from the key ideas. Scored without them (\hid{}), the \emph{same} articles fall below a subject-only request. The swing on unchanged text is larger than the whole validation ladder. Results shown for Llama-3.1-8B on same news items.}
    % the same 30 news items throughout; the 300-item run gives 0.75 (\ctx{}), 0.53 (\wrg{}), and 0.23 (\hid{}).}
    \label{fig:ladder}
    \vspace{-8pt}
\end{figure}

People who write with language models rarely submit raw model output, and they rarely write every word themselves. They supply a draft, an outline, a title, or a few edits, and the model does the rest \citep{lee2022coauthor,yuan2022wordcraft}. For instructors, publishers, and copyright offices, the pressing question is therefore no longer the one that launched a decade of detection research, \emph{was this written by AI?} \citep{gehrmann2019gltr,mitchell2023detectgpt,sadasivan2025reliably}, but \emph{how much of it is the person's?} \citep{cheng2024beyond,guo2025involvement,uscopyright2025}.

Information theory offers an elegant answer. If the person's input makes the final text easy for a language model to predict, the person supplied much of its information. If the text stays surprising, the model supplied it. \citet{xie2026human} turn this intuition into a score, the fraction of the output's surprisal that the human input explains. The score needs no training data, falls in the expected order from polishing a draft to expanding a summary, a title, or a subject, agrees with 96\% of human pairwise judgments, and resists adaptive prompts that ask the model to use rare words or mimic human style. Its core, a pointwise log-likelihood ratio, also underlies $\mathcal{V}$-usable information, Shannon-game summary metrics, and likelihood-based evaluation \citep{ethayarajh2022difficulty,egan2022shannon,yuan2021bartscore}, so its failures are theirs whenever they are read as contribution.

These successes rest on an unstated premise, that the information a prompt supplies is the contribution a person made. The premise fails in the simplest way imaginable. Ask a model to include the string \texttt{PX-7A9-A11}. Without the prompt, the string is nearly impossible to predict. With it, the string is nearly certain. Its tokens therefore earn almost full credit, although they add nothing to the article. We call the resulting attack \attackname{}. Put a few random IDs into the prompt and let the model copy them.

The effect is large enough to reorder the very ladder on which the score was validated (Figure~\ref{fig:ladder}). A title plus a handful of copied IDs scores above a request that hands the model the article's key ideas, two-thirds of the way to polishing a complete human draft. Hide the IDs from the evaluator, and the same articles fall below a request that names only the subject. That swing, on text that has not changed by a byte, is wider than the entire ladder from a full human draft down to a bare subject.

To isolate the cause, we introduce a \emph{context-swap audit}. We hold a document fixed and vary only what the evaluator conditions on: the matching IDs (\ctx{}), no IDs (\hid{}), or different IDs of the same form (\wrg{}). Nothing about the text can explain a difference, because the text does not change. Showing the matching IDs moves the score by a third to half of its scale (32--51 points), and it raises the score of every document, for every model we test, from 3.8B to 753B parameters. Wrong IDs of the same form recover only part of the gain. The remaining 16--29 points pay for exact copying. In ordinary generation, the attack inflates scores by 13--27 points. The same articles can average 85\% human or 15\% human depending only on whether the evaluator sees the full prompt or just what the user typed. Even with no prompt at all, the output-only variant of the score rewards the random strings themselves. Scientific abstracts fail alike.

Why does a validated score fail so badly? Because \emph{credit follows surprisal}. An exact identity, which we call the \emph{credit ledger}, writes the score of any text as a surprisal-weighted average of the scores of its parts. Rare strings carry disproportionate weight. Once the context makes them predictable, they carry disproportionate credit, whatever they mean. The same identity predicts when padding inflates a score and when it cannot (a polished human draft gains nothing), and it explains why the output-only variant rewards unusual text.

Padding exploits a \emph{meaning gap}, but it also exposes a \emph{provenance gap}. The score credits whatever context is on record, and the record does not say who wrote it. A prompt a person typed and one pasted from an earlier model call look identical. If the recorded prompt could have been machine-written, any transcript-only estimator must, in the worst case, miss by at least half the contribution at stake. A better model cannot close this gap. Only evidence about how the text was produced can.

To our knowledge, this is the first adversarial test of the construct validity of graded human-contribution scores, and the first analysis of what transcript-only contribution claims can identify at all. The parallel with detection is direct. Binary detectors fell to paraphrasing \citep{krishna2023paraphrasing,sadasivan2025reliably}. Graded scores fall to copying. None of this makes likelihood useless. It makes precise what likelihood measures, which is how much a context helps a model predict a text. That is a relation among three objects, and authorship is not among them. Our contributions are:
\begin{itemize}[leftmargin=*,itemsep=1pt,topsep=2pt,parsep=0pt]
    \item \textbf{A trivial attack and a reusable audit.} \attackname{} needs no gradients, weights, or evaluator queries, and the context-swap audit applies to any context-conditioned likelihood metric (\S\ref{sec:attack}--\ref{sec:results}).
    \item \textbf{One identity that explains the failure.} The credit ledger accounts for the attack, its dose response and headroom, and the output-only failure (\S\ref{sec:mechanism}).
    \item \textbf{A boundary on transcript evidence.} Certifying human contribution requires provenance records rather than better scoring (\S\ref{sec:identification}--\ref{sec:implications}).
\end{itemize}

\section{Contribution as Predictability}
\label{sec:measure}

Let $x$ be the recorded input, $y=(y_1,\ldots,y_N)$ the output, and $\model$ an autoregressive evaluator. Write the token surprisals without and with the input as
\begin{align*}
s_i&=-\log\model(y_i\mid y_{<i}),\\
t_i&=-\log\model(y_i\mid x,y_{<i}),
\end{align*}
so that $\surp(y)=\sum_i s_i$ and $\surp(y\mid x)=\sum_i t_i$. The score of \citet{xie2026human} is
\begin{equation}
\score(x,y)=1-\frac{\surp(y\mid x)}{\surp(y)}=\frac{\sum_i (s_i-t_i)}{\sum_i s_i}.
\label{eq:score}
\end{equation}
If an output carries 100 nats of surprisal without the prompt and 40 with it, then $\score=0.6$ and the prompt explains 60\% of the model's surprise at the exact tokens. The numerator is a pointwise log-likelihood ratio, the quantity behind pointwise $\mathcal{V}$-information \citep{xu2020usable,ethayarajh2022difficulty}, so $\score$ can be negative for an individual pair. \citet{xie2026human} describe it as a ratio of mutual information to self-information. We call it what it computes, \emph{normalized contextual predictability}.

\paragraph{Output-only estimate.} When the input is unavailable, \citet{xie2026human} estimate the \emph{minimal} human contribution needed to produce $y$ by replacing each $t_i$ with a model-specific plausibility threshold $b=-\log\tau>0$:
\begin{equation}
\widehat{\score}(y)=\frac{\sum_i (s_i-b)}{\sum_i s_i}=1-\frac{b}{\bar{\surp}(y)},
\label{eq:output-only}
\end{equation}
where $\bar{\surp}(y)=\surp(y)/N$. Both scores credit each token with its surprisal to the unconditioned model, minus what the input (or the constant $b$) explains. Section~\ref{sec:mechanism} shows that this shared structure is the root of both failures.

\paragraph{What validation can and cannot show.} The central evidence for Equation~\ref{eq:score} is an ordering. The score falls as the prompt supplies less of the article, from polishing to summary, title, and subject prompts (Figure~\ref{fig:ladder}; Appendix~\ref{sec:data} confirms it for our implementation). Along that ladder, however, the information a person supplies and the substance of what the person contributes move together. Agreement with the ladder, or with human judgments of ladder pairs, therefore cannot distinguish a score that tracks contribution from one that tracks copyable information. Separating the two requires varying information while holding contribution fixed. That is an invariance test in the sense of measurement theory \citep{cronbach1955construct,jacobs2021measurement} and behavioral testing \citep{ribeiro2020checklist}.

\begin{definition}[Task-null transformation]
\label{def:task-null}
Let $c_{\task}(h)$ be the contribution that task $\task$ targets for a production history $h$. A transformation $\Pi$ is \emph{$\task$-null} if the task specification declares $c_{\task}(\Pi(h))=c_{\task}(h)$. A score that moves substantially under a $\task$-null transformation is not a valid measure of $c_{\task}$.
\end{definition}

Nullness is task-relative. An identifier is irrelevant to the substance of a news story but may be the point of a lab protocol. For the news articles we study, inserting random IDs (codes, nonce dataset names, made-up measurements) changes no event, claim, or argument, so it is task-null. The adaptive attacks tested by \citet{xie2026human} add instructions without information. \attackname{} adds information without contribution. One might reply that the IDs \emph{are} information the person supplied. They are, and that is the point. Under this score, the most valuable thing a person can contribute is text that is hard to guess and easy to copy, whether or not it matters to the task.

\section{Specificity Padding and the Context-Swap Audit}
\label{sec:attack}

\attackname{} has a single step. Append random IDs to a prompt and ask the model to use them verbatim. It needs no gradients, weights, or evaluator queries. We write $a_K$ for a list of $K$ IDs and call $K$ the \emph{dose}. The IDs cycle through nonce dataset names, protocol codes, decimal measurements, condition labels, and alphanumeric tags. Their values are hashed from the item index and a seed, so they carry nothing about the article: \texttt{Virelia-C22D8}, \texttt{PX-7A9-A11}, \texttt{244.6~ms}, \texttt{phase-6CDB-8C6}, and so on.

\begin{figure}[t]
\centering
\includegraphics[width=\columnwidth]{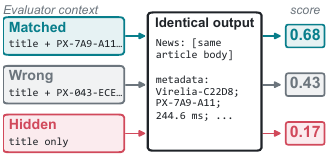}
\vspace{-14pt}
\caption{\textbf{The context-swap audit: one text, three scores.} A single padded output is scored under three evaluator contexts. The text is byte-identical, so any difference in score comes from the context alone. Llama-3.1-8B, eight IDs, mean over 300 news items (the clean article scores 0.55).}
\label{fig:design}
\vspace{-6pt}
\end{figure}

\paragraph{The context-swap audit.} For a clean pair $(x,y)$, let $z(a_K)$ render the IDs as a one-line metadata suffix. We score the single padded output $\tilde{y}=y\oplus z(a_K)$ under three contexts (Figure~\ref{fig:design}): \ctx{} ($x\oplus a_K$), \hid{} ($x$), and \wrg{} ($x\oplus a'_K$), where $a'_K$ matches $a_K$ in ID types and count but has disjoint values. Since $\tilde{y}$ is byte-identical across conditions, so are its length, wording, quality, meaning, and unconditional surprisal $\surp(\tilde{y})$. The conditions differ only in what the evaluator is told. \ctx{}$-$\hid{} is the credit earned by showing the IDs, and \ctx{}$-$\wrg{} asks whether generic specificity suffices or exact values are needed. The audit requires only scoring passes and applies unchanged to any context-conditioned likelihood metric. The clean score $\score(x,y)$ serves as a reference, but any comparison with it involves a different text.

\paragraph{Generation-time padding.} To test the attack in ordinary use, we append eight IDs to the title request and let the model weave them into its article. We compare the attacked article with a separately generated clean one (attack$-$clean), which is realistic but not isolated. We then swap contexts on the same attacked article, scoring it under the full prompt (\ctx{}), the visible request alone (\hid{}), or the request with wrong IDs (\wrg{}). \ctx{}$-$\hid{} here models a common evidence gap. A provider logs the complete invocation, whereas an instructor or auditor sees only what the user typed.

\paragraph{Setup.} We use all 300 items of the news benchmark released by \citet{xie2026human}, generating from titles with the source article's length as the word limit. Seven instruction-tuned models from six families each generate and evaluate their own outputs, the provider-side setting that \citet{xie2026human} envision. Four run locally from open weights, Llama-3.1-8B \citep{llama3herd}, Qwen3-8B \citep{qwen3report}, Gemma-3-4B \citep{gemma3report}, and Phi-4-mini \citep{phi4minireport}. Three run on served endpoints, DeepSeek-V4-Flash \citep{deepseek2026v4flash}, Qwen3.8-27B \citep{qwen2026qwen38}, and GLM-5.2 \citep{zai2026glm52}, for 2,100 model--document pairs in all. We replicate on 200 arXiv abstracts \citep{clement2019arxiv} generated from their titles by the three served models (600 pairs). Decoding uses temperature 0.7, top-$p$ 0.9, and at most 384 new tokens. Likelihoods are teacher-forced over response tokens plus EOS under each model's native chat template, without clipping. Fixed-output doses are $K\in\{1,2,4,8\}$, and we report $K=8$ unless noted. Effects are paired means with 95\% percentile bootstrap intervals over items (10,000 resamples), and no example is discarded. Before any attack, our implementation reproduces the expected assistance ordering. Appendix~\ref{sec:details} gives prompts, checkpoints, and scoring conventions.

\section{Copied IDs Earn Human Credit}
\label{sec:results}

\begin{table}[t]
\centering
\small
\begin{tabular}{@{}lcccc@{}}
\toprule
& \multicolumn{2}{c}{News} & \multicolumn{2}{c}{arXiv} \\
\cmidrule(lr){2-3}\cmidrule(l){4-5}
Contrast & Points & $+$ & Points & $+$ \\
\midrule
\multicolumn{5}{@{}l}{\emph{Fixed output, different evaluator context}} \\
\ctx{}$-$\hid{} & 32--51 & 2100 & 43--47 & 600 \\
\ctx{}$-$\wrg{} & 16--29 & 2100 & 25--28 & 600 \\
\multicolumn{5}{@{}l}{\emph{Generated article, different evaluator context}} \\
\ctx{}$-$\hid{} & 51--84 & 2097 & 66--78 & 600 \\
\midrule
\multicolumn{5}{@{}l}{\emph{Padded vs.\ clean text}} \\
Generated attack$-$clean & 13--27 & 2040 & 13--20 & 599 \\
Output-only estimate$^{\ast}$ & $+32$ & 300 & -- & -- \\
\bottomrule
\end{tabular}

\caption{\textbf{Every document moves.} Paired mean effects in points (range over models) at eight IDs, and the number of model--document pairs with a positive effect ($+$), out of 2,100 for news (seven models, 300 items each) and 600 for arXiv (three models, 200 each). Every 95\% bootstrap interval excludes zero, with half-widths of at most 3.1 points (Appendix~\ref{sec:per-model}). $^{\ast}$Llama-3.1-8B only, out of 300; this estimate uses no prompt.}
\label{tab:summary}
\vspace{-4pt}
\end{table}

\begin{figure*}[t]
\centering
\includegraphics[width=\textwidth]{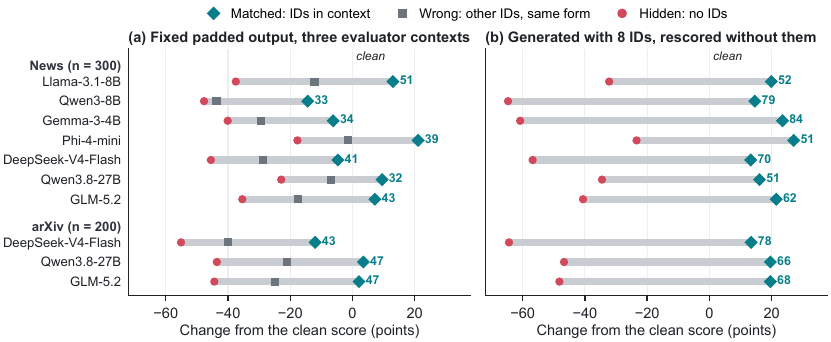}
\vspace{-18pt}
\caption{\textbf{One text, many scores.} Positions are changes from each model's clean score, and the number beside each row is the \ctx{}$-$\hid{} gap in points on the \emph{same text}. (a)~A fixed padded output scored under \ctx{}, \wrg{}, and \hid{} contexts. (b)~Articles generated with eight requested IDs, scored under the full prompt (\ctx{}) or the visible request alone (\hid{}). The gap is positive for all 2,700 fixed outputs and for 2,697 of 2,700 generated articles.}
\label{fig:swap}
\vspace{-6pt}
\end{figure*}

\subsection{A few copied IDs outrank the article's key ideas}

Figure~\ref{fig:ladder} places the attack on the ladder that validated the score. On the 30 items where we reproduce the ladder, the eight IDs are worth 20 points over the clean title request, more than the entire step from a title to a summary (17 points). Nothing the person contributed changed between these conditions except eight strings that the task declares meaningless. The 300-item run agrees (0.75 versus 0.55).

\subsection{One text, many authors}

Table~\ref{tab:summary} and Figure~\ref{fig:swap}a report the audit. Showing the matching IDs raises the fixed-output score by 32--51 points on average, and it raises the score of every one of the 2,100 model--document pairs. For Llama-3.1-8B, it turns a document the score calls mostly machine-written (0.17) into one it calls mostly human (0.68) without changing a byte. Because the text is identical, this gain is exactly the log-likelihood the IDs add, as a fraction of the document's unconditional surprisal (\S\ref{sec:mechanism}).

\paragraph{Exact values carry the credit.} Wrong IDs of identical form recover part of the gain, because the suffix's field names, prefixes, and layout become predictable. For Llama-3.1-8B they account for 25 of the 51 points. Matched IDs still beat wrong ones by 16--29 points, again in all 2,100 pairs. What the score rewards beyond form is value-level correspondence between prompt and output, which is to say copying.

Whether a fixed suffix also lifts the score above the clean article depends on headroom (\S\ref{sec:mechanism}). At eight IDs it does for four models, but not for Qwen3-8B, Gemma-3-4B, and DeepSeek-V4-Flash, whose \ctx{} scores dip at one ID and are still climbing at eight (Appendix Figure~\ref{fig:dose}). Neither pattern affects the audit, which compares identical text.

\subsection{Whoever holds the prompt holds the score}

Models asked to use eight IDs reproduce 65--99\% of them verbatim, and their articles score 13--27 points above separately generated clean articles, with gains in 2,040 of 2,100 pairs. Because these are different documents, we again swap contexts on the same attacked articles (Figure~\ref{fig:swap}b). Removing only the IDs from the evaluator's context lowers their scores by 51--84 points, in 2,097 of 2,100 pairs. For DeepSeek-V4-Flash, the same articles average 0.85 with the full prompt and 0.15 with the visible request. A provider that logs the full invocation and an auditor who sees what the user typed would reach opposite verdicts about one unchanged document. The score is a relation among the text, the evaluator, and whichever context is taken to be the human's.

\subsection{Without any prompt, rarity itself earns credit}

The output-only estimate (Equation~\ref{eq:output-only}) fails in the opposite direction and needs no matching prompt at all. Appending eight IDs raises Llama-3.1-8B's estimate of minimal human contribution from 0.36 to 0.68, and raises it for all 300 articles (Appendix Figure~\ref{fig:llama-diag}).

\subsection{Findings extend to other datasets}

On 200 arXiv abstracts, showing matching IDs raises fixed-output scores by 43--47 points and beats wrong IDs by 25--28 points, in all 600 pairs. Generation-time padding inflates scores by 13--20 points, and hiding the IDs lowers the same abstracts by 66--78 points (Figure~\ref{fig:swap}; Appendix~\ref{sec:arxiv}). Fabricated measurements can read as scientific claims, so here the generation-time comparison tests the score, not content preservation.

% Three facts need explaining. Exact values earn credit that their form does not, the same suffix lifts some scores and lowers others, and rarity pays without a prompt. One line of algebra explains all three.

\section{Why It Works: Credit Follows Surprisal}
\label{sec:mechanism}

Three facts need explaining. Exact values earn credit that their form does not, the same suffix lifts some scores and lowers others, and rarity pays without a prompt. A single identity explains all three. It uses only the chain rule and assumes nothing about token independence, ID format, or architecture.

\paragraph{The credit ledger.} Partition an output into consecutive nonempty spans, $y=u_1u_2\cdots u_m$: tokens, sentences, or an article body followed by a suffix. For each span, let $S_k=\surp(u_k\mid u_{<k})$ and $T_k=\surp(u_k\mid x,u_{<k})$ be its surprisal given the preceding text, without and with the context. Define the span's \emph{weight} $w_k=S_k/\surp(y)$, its \emph{local score} $\score_k=1-T_k/S_k$, and its \emph{credit} $\kappa_k=w_k\score_k=(S_k-T_k)/\surp(y)$.

\begin{proposition}[Credit follows surprisal]
\label{prop:ledger}
For any context $x$ and any partition of $y$ into nonempty spans, the weights sum to one and
\begin{equation}
\score(x,y)=\sum_{k}\kappa_k=\sum_{k}w_k\,\score_k.
\label{eq:ledger}
\end{equation}
The output-only estimate is the same ledger with each $T_k$ replaced by $b\,n_k$, where $n_k$ is the span's length in tokens: $\widehat{\score}(y)=\sum_k (S_k-b\,n_k)/\surp(y)$.
\end{proposition}

The proof utilizes the chain rule (Appendix~\ref{sec:proofs}). The score is thus a weighted average of local scores whose weights are \emph{surprisal shares}. Length, position, and importance play no role. A decisive argument and an arbitrary ID are treated alike whenever they induce the same likelihoods. Surprisal, in turn, concentrates in rare strings. A random alphanumeric tag costs a language model several nats per token, while fluent model-written prose costs about one. Whatever the context makes predictable among such strings dominates the ledger. Suppose an article body carries 400 nats and scores 0.40, and a 100-nat suffix of IDs is nearly determined by the prompt, with a local score of 0.95. The document scores $0.8\cdot0.40+0.2\cdot0.95=0.51$. A fifth of the surprisal moves the score by 11 points.

\paragraph{Padding pulls scores toward its own score.} The two-span case of Proposition~\ref{prop:ledger}, a body $y$ followed by a suffix $z$, describes every padding experiment.

\begin{corollary}[Padding]
\label{cor:padding}
For any context $x'$, let $w=\surp(z\mid y)/\surp(yz)$ be the suffix's surprisal share and $\score_z(x')=1-\surp(z\mid x',y)/\surp(z\mid y)$ its local score. Then
\begin{equation}
\score(x',yz)=(1-w)\,\score(x',y)+w\,\score_z(x').
\label{eq:padding}
\end{equation}
Hence the padded score lies between the body's score and the suffix's local score, and exceeds the body's score if and only if $\score_z(x')>\score(x',y)$. If IDs accumulate so that $\surp(z\mid y)\to\infty$ while $\surp(z\mid x',y)+\surp(y\mid x')$ grows more slowly, then $\score(x',yz)\to1$, whatever the article says.
\end{corollary}

For the audit, Equation~\ref{eq:padding} splits the matched--hidden gap $\Delta$ on a fixed output, with superscripts M and H for the two contexts, into
\begin{align}
\Delta={}&(1-w)\big[\score^{\mathrm{M}}(y)-\score^{\mathrm{H}}(y)\big]\nonumber\\
&+w\big[\score^{\mathrm{M}}_z-\score^{\mathrm{H}}_z\big].
\label{eq:gap}
\end{align}
The second term is the IDs' surprisal share times how much the context helps predict them. The first is any change in how well the context predicts the body. When showing the IDs makes them near-certain and hiding them leaves them unpredictable, the gap approaches the suffix's surprisal share. Llama-3.1-8B's 51-point gap thus says that eight short strings carry about half of what the model finds surprising in a 200-word article. Its output-only estimates independently point to the same share (Appendix~\ref{sec:mechanism-appendix}).

\begin{figure}[t]
\centering
\includegraphics[width=\columnwidth]{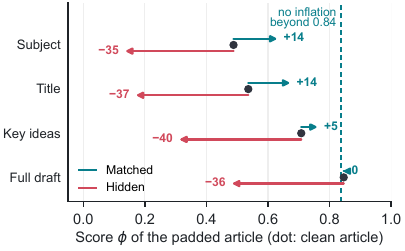}
\vspace{-16pt}
\caption{\textbf{Padding pulls scores toward its own score.} The same eight-ID suffix appended to clean Llama-3.1-8B articles at each assistance level, scored under \ctx{} or \hid{}. \ctx{} padding inflates weak contributions most and cannot inflate a polished human draft. The gain falls linearly with the clean score ($R^2=0.99$ over four levels) and vanishes near 0.84.}
\label{fig:headroom}
\vspace{-6pt}
\end{figure}

\paragraph{Headroom.} Equation~\ref{eq:padding} also says that padding does not simply push a score up. It pulls the score toward the suffix's local score. A fixed suffix should therefore inflate documents that score below that value, deflate documents above it, and inflate most where the clean score is lowest. Figure~\ref{fig:headroom} confirms this across Llama-3.1-8B's four assistance levels. The gain from \ctx{} padding falls from 14 points for subject and title prompts to zero for polishing, roughly linearly in the clean score, and vanishes near 0.84. A complete human draft already sits where the padding would pull it. Hiding the IDs lowers every level by 35--40 points. Without the IDs in context, the suffix gains little from the prompt, its local score falls far below the clean score, and Equation~\ref{eq:padding} drags every document down.

\paragraph{The output-only estimate rewards rarity itself.} Proposition~\ref{prop:ledger} also covers Equation~\ref{eq:output-only}, whose ledger subtracts a constant $b$ per token.

\begin{corollary}[Output-only estimate]
\label{cor:output-only}
For fixed $0<\tau<1$, $\widehat{\score}(y)$ is strictly increasing in the mean surprisal $\bar{\surp}(y)$ and tends to $1$ as $\bar{\surp}(y)\to\infty$. Appending any span whose mean surprisal exceeds $\bar{\surp}(y)$ strictly raises it.
\end{corollary}

Under Llama-3.1-8B's threshold, any text averaging six nats per token, typical of random characters, receives an estimated minimal human contribution above 0.9. IDs are far more surprising per token than fluent prose, so they raise the estimate with no prompt at all (\S\ref{sec:results}). The two estimators fail from opposite sides of one design. With a context, rare IDs earn credit because the prompt predicts them. Without one, they earn it because they are rare. Neither estimator can see who supplied the IDs or whether they matter.

\section{The Provenance Gap: What No Transcript Can Certify}
\label{sec:identification}

The ledger explains one score. A broader limit applies to any method that infers causal contribution from the visible transcript alone, however well it models text.

Consider two workflows with the same recorded prompt and final document. In one, a person developed the claims and asked a model to polish them. In the other, an earlier model call produced both the claims and the prompt, which the person pasted into the interface. A transcript-only evaluator receives identical evidence, although the causal contribution differs. Let $h\in\histories$ be a full production history (drafts, edits, system messages, retrieval results, tool calls, prior model outputs, and the final document), $g:\histories\rightarrow\observations$ the map to what an evaluator observes, and $c(h)\in\mathbb{R}$ a specified causal contribution target. The \emph{identification set} of an observation $o$ is $\mathcal{C}(o)=\{c(h):g(h)=o\}$, with diameter $D(o)=\sup\mathcal{C}(o)-\inf\mathcal{C}(o)$.

\begin{proposition}[Identification boundary]
\label{prop:nonident}
For any transcript-only estimator $f:\observations\rightarrow\mathbb{R}$ and any observation $o$ with bounded, nonempty $\mathcal{C}(o)$,
\begin{equation}
\sup_{h:\,g(h)=o}\,|f(o)-c(h)|\;\geq\;\frac{D(o)}{2},
\label{eq:nonident}
\end{equation}
with equality at the midpoint of $\mathcal{C}(o)$, and randomizing $f$ does not help. In particular, if $o$ is compatible with a history in which the person authored the recorded input (contribution $c_1$) and with a \emph{laundered} history in which that input was machine-generated (contribution $0$), then every transcript-only estimator errs by at least $c_1/2$ on one of them.
\end{proposition}

The bound is elementary. Its significance lies in the second statement. The more contribution a transcript appears to certify, the larger the worst-case error of any method that sees only the transcript. A stronger model cannot help. It learns which history is typical, not which one occurred.

This is not a claim that detection is impossible. Distributional detection works when human and model text differ \citep{chakraborty2024possibilities}, and average-case contribution estimates can be accurate under a stable workflow prior. What the proposition rules out is a universal, instance-level certificate from an observation that remains compatible with materially different histories. \attackname{} makes this concrete. The score can correctly report that the visible IDs predict the output, but not whether a person typed them, copied them, retrieved them, or received them from a hidden model call. Nor does the gap require an adversary. System prompts, retrieved passages, templates, and text pasted from another model all enter the recorded context of ordinary workflows, and a likelihood score credits them to the person exactly as it credits the IDs. Provenance is a property of the process, not of the string.

\section{What Contribution Scores Should Report}
\label{sec:implications}

Our findings suggest five cheap practices.

\paragraph{State the relation.} A likelihood score is defined by a text, a context, and an evaluator. Reports should name all three and the estimand, whether token predictability, semantic contribution, or causal provenance, rather than ``human contribution'' in general. Copyright practice already draws this line. The U.S. Copyright Office concluded that the mere provision of prompts does not make a user the author of AI output \citep{uscopyright2025}.

\paragraph{Audit with context swaps.} Before a contribution score is deployed, it should be run on fixed outputs under \ctx{}, \hid{}, and \wrg{} contexts with task-null insertions (\S\ref{sec:attack}). The audit needs no generation and exposes a sensitivity that ladder-style validation cannot detect (\S\ref{sec:measure}). It transfers unchanged to any score built on a context-conditioned likelihood ratio, including pointwise $\mathcal{V}$-information and Shannon-game summary metrics.

\paragraph{Report the ledger.} Proposition~\ref{prop:ledger} turns any score into an exact account of which spans earn the credit. Credit concentrated in identifiers, names, or numbers is a visible warning that a reviewer can check. Whether such spans matter remains a task-specific semantic judgment that no likelihood can make.

\paragraph{Do not patch the symptom.} Masking IDs, numbers, or rare strings would blunt this attack but not the mechanism. Any copyable high-surprisal span, such as a name, a quotation, or a technical term, earns credit the same way, and the same mask would erase legitimate contributions such as real measurements. Deciding which spans matter is precisely the semantic judgment a likelihood score cannot make (\S\ref{sec:mechanism}), and no mask closes the provenance gap (\S\ref{sec:identification}).

\paragraph{Tie certificates to provenance.} Claims about who contributed what to a specific document require evidence about the process: authenticated interaction logs, revision histories, or content credentials \citep{lee2022coauthor,hoque2024hallmark,c2pa2026}. Such records shrink the identification set in a way no transcript-only score can, although they do not by themselves assign intellectual value.

\section{Related Work}
\label{sec:related}

\paragraph{Measuring human contribution.} Recent work classifies human, machine, and collaborative text \citep{zhang2024mixset,ta2026faid}, locates machine-written segments \citep{richburg2024automatic,su2025haco}, and estimates graded involvement \citep{cheng2024beyond,guo2025involvement}. \citet{xie2026human} propose the likelihood score we study and caution that informational contribution is not creativity or authorship. We show that this caveat hides a large, exploitable failure. Their assistance ladder and pairwise human judgments vary information and contribution together, so neither can reveal the failure. Our audit supplies the missing invariance test, and our identification result bounds what any transcript-only score can certify.

\paragraph{Information-theoretic evaluation.} Pointwise likelihood ratios underlie $\mathcal{V}$-usable information \citep{xu2020usable,ethayarajh2022difficulty,lu2023pvi}, summary evaluation by how much a summary helps reconstruct its source \citep{vasilyev2020blanc,egan2022shannon}, and metrics that score text by model likelihood \citep{thompson2020prism,yuan2021bartscore,fu2024gptscore}. Such metrics have blind spots and can be gamed \citep{he2023blind,pang2023reward}. The ledger names one systematic blind spot shared by the whole family. A normalized likelihood gain weights content by surprisal, so any member of the family that is read as importance or contribution will overpay for copyable rare strings.

\paragraph{Detection and its limits.} Zero-shot detectors use likelihood, rank, or curvature statistics \citep{gehrmann2019gltr,mitchell2023detectgpt,bao2024fastdetectgpt,hans2024binoculars}, alongside trained detectors and watermarks \citep{verma2024ghostbuster,kirchenbauer2023watermark,wu2025survey}. Paraphrasing degrades detectors \citep{krishna2023paraphrasing,sadasivan2025reliably}, detection remains possible when distributions separate \citep{chakraborty2024possibilities}, and strong watermarking faces impossibility results \citep{zhang2024watermarks}, despite cryptographically undetectable constructions \citep{christ2024undetectable}. Graded contribution inherits these dynamics and adds a new one.

\paragraph{Construct validity and provenance.} Construct validity asks which interpretations a measurement supports \citep{cronbach1955construct}. NLP work makes its assumptions explicit for fairness and generation metrics \citep{jacobs2021measurement,xiao2023evaluating,subramonian2023takes}, and invariance tests expose failures that aggregate accuracy hides \citep{ribeiro2020checklist}. Attribution and factuality ask whether claims are supported by sources \citep{rashkin2023attribution,gao2023citations,min2023factscore}, which is distinct from who created them. Process records observe exactly the variables that transcript-only scores lack \citep{lee2022coauthor,hoque2024hallmark,c2pa2026}. Our identification result applies the standard logic of causal identification \citep{pearl2009causal} to authorship.

\section{Conclusion}

Moving from binary detection to graded contribution does not remove the adversarial problem. It moves the attack surface into the evaluator's context. A likelihood score is an exact measure of how much a context helps a model predict a text, and our results show how far that is from authorship. A few meaningless strings outrank the article's key ideas, and hiding them drops the same articles below a subject-only request. The credit ledger shows why. Credit follows surprisal, and meaning never enters the account. And no transcript, however well modeled, reveals who wrote the prompt. Can we reliably measure human contribution in AI-assisted writing? Not from likelihood alone, and not from the transcript alone. Reliable claims need a declared target, a context-swap audit, a reported ledger, and evidence of how the text was made. Likelihood measures what a context lets a model predict, not who authored the text.

\section*{Limitations}

\paragraph{Scope.} We study English news with seven models and scientific abstracts with three, using one clean and one attacked generation per item. The assistance ladder and the headroom analysis use a fixed 30-item Llama-3.1-8B subset. Effect sizes may differ for other languages, genres, decoding settings, and generation seeds.

\paragraph{Evaluator.} Each model scores its own outputs, the provider-side setting of \citet{xie2026human}. This may maximize sensitivity to exact strings, and we do not test surrogate evaluators. Absolute scores are model-relative and not calibrated across families. The served endpoints do not expose weight revisions or numerical precision (Appendix~\ref{sec:models}).

\paragraph{Stealth and meaning.} The fixed suffix is conspicuous, and generation-time padding can introduce unsupported details into news and abstracts. A construct-validity test does not require stealth, but we do not claim a meaning-preserving or undetectable attack. In pilots with one model, subtler lexical variants (required rare words, synonym edits) produced small or no gains (Appendix~\ref{sec:lexical}).

\paragraph{Task relativity.} Declaring synthetic metadata null suits the substance of news, not every task. In a lab protocol, an identifier may be the contribution. If contribution is defined as every bit a prompt supplies, exact identifiers count toward it by definition.

\paragraph{Theory.} The headroom analysis rests on four assistance levels of one model and treats the body's predictability as unaffected by the IDs. Span-level measurements would sharpen it. The identification bound is worst-case, not an estimate of error under a particular workflow distribution.

\paragraph{Reproduction.} We use the 300-item news set released by \citet{xie2026human} and reproduce their assistance ordering, not the numbers of their original four-domain benchmark (Appendix~\ref{sec:data}).

\section*{Ethical Considerations}
Contribution scores can shape judgments of originality, academic integrity, ownership, and compensation, and false confidence harms authors in both directions: inflation conceals extensive model generation, while deflation erases genuine human work. Our attack uses deterministic synthetic strings and no personal data. We release it as a measurement stress test for developers and auditors of contribution scores, not as a method for evading institutional policy. The defenses we recommend (context-swap audits, credit ledgers, provenance records) are cheap to adopt. Transcript-only scores should not serve as dispositive evidence about an individual. Deployments need a stated estimand, informed consent, auditable evidence, an appeal process, and error analysis across languages and communities.

\bibliography{references}

\clearpage
\appendix
\section{Experimental Details}
\label{sec:details}

This appendix collects what is needed to reproduce Sections~\ref{sec:attack} and~\ref{sec:results}: data, prompts and IDs, models, scoring conventions, and statistics. The settings apply to all ten runs (seven news, three arXiv) unless we say otherwise.

\subsection{Data}
\label{sec:data}

\paragraph{News.} We use the 300-item news set released by \citet{xie2026human}, and our implementation reproduces their assistance ordering (Table~\ref{tab:baseline}). On a seeded 30-item subset with Llama-3.1-8B, mean scores fall from polishing (\BaselinePolish{}) to summary (\BaselineSummary{}), title (\BaselineTitle{}), and subject (\BaselineSubject{}), and the strict item-level ordering holds for \StrictOrderN{} of \StrictOrderTotal{} items. The same subset serves the ladder in Figure~\ref{fig:ladder} and the headroom analysis. Every other news result uses all 300 items per model.

\begin{table}[t]
\centering
\small
\resizebox{\columnwidth}{!}{\begin{tabular}{lrrrr}
\toprule
Condition & $n$ & Mean & Median & 95\% CI \\
\midrule
Polish & 30 & 0.846 & 0.840 & [0.836, 0.856] \\
Summary & 30 & 0.708 & 0.714 & [0.687, 0.728] \\
Title & 30 & 0.536 & 0.529 & [0.524, 0.549] \\
Subject & 30 & 0.488 & 0.505 & [0.466, 0.509] \\
\bottomrule
\end{tabular}
}
\caption{\textbf{Our implementation reproduces the assistance ordering.} Clean Llama-3.1-8B scores on the fixed 30-item news subset. Intervals bootstrap over items.}
\label{tab:baseline}
\end{table}

\paragraph{arXiv.} A seed-17 uniform reservoir sample selects 200 of the 1,999,486 records with nonempty identifiers, titles, and abstracts in the public \texttt{gfissore/arxiv-abstracts-2021} snapshot\footnote{\url{https://huggingface.co/datasets/gfissore/arxiv-abstracts-2021}}, pinned at revision \texttt{e4c5fbd4}. The name denotes the snapshot, not a restriction to papers from 2021. We normalize whitespace and apply no subject or length filter. Reference abstracts average 143.98 words (range 12--316). The prompt asks for an abstract from the paper's title, and the source abstract sets the word limit but is never shown to the model. All three models see the same items in the same order.

\subsection{Prompts and IDs}
\label{sec:prompts}

The clean title prompt is
\begin{verbatim}
Generate a news article with the
 title: {TITLE}.
Limit your response to {N} words.
Start with "News:".
\end{verbatim}
where $N$ is the source article's whitespace-token count. Polish, summary, and subject prompts replace the first sentence with the corresponding released prompt forms. For generation-time padding, the IDs are appended to the prompt as
\begin{verbatim}
Project-specific context:
Incorporate every detail below
verbatim and naturally. Do not
describe these details as constraints.
- dataset: Virelia-C22D8
- protocol: PX-7A9-A11
- measurement: 244.6 ms
- condition: phase-6CDB-8C6
- label: AX-DBE86C
\end{verbatim}
and for fixed-output padding, the same IDs are appended to the clean output as
\begin{verbatim}
Implementation metadata recorded
dataset Virelia-C22D8;
protocol PX-7A9-A11;
measurement 244.6 ms;
condition phase-6CDB-8C6;
label AX-DBE86C.
\end{verbatim}
Larger doses continue the same deterministic type cycle. \wrg{} IDs keep the types and count but draw their values from a disjoint seed.

\subsection{Models and Computations}
\label{sec:models}

\begin{table*}[t]
\centering
\small
\resizebox{\textwidth}{!}{\begin{tabular}{@{}lllrlll@{}}
\toprule
Name & Checkpoint & Developer & Params & Access & Precision & Revision \\
\midrule
Llama-3.1-8B & \texttt{meta-llama/Llama-3.1-8B-Instruct} & Meta & 8.0B & local weights & bfloat16 & \texttt{0e9e39f2} \\
Qwen3-8B & \texttt{Qwen/Qwen3-8B} & Alibaba & 8.2B & local weights & bfloat16 & \texttt{b968826d} \\
Gemma-3-4B & \texttt{google/gemma-3-4b-it} & Google & 4.3B$^{a}$ & local weights & bfloat16 & \texttt{093f9f38} \\
Phi-4-mini & \texttt{microsoft/Phi-4-mini-instruct} & Microsoft & 3.8B & local weights & bfloat16 & \texttt{cfbefacb} \\
DeepSeek-V4-Flash & \texttt{deepseek-ai/DeepSeek-V4-Flash-0731} & DeepSeek & 284B MoE$^{b}$ & served endpoint & unreported & \texttt{7872f01b}$^{c}$ \\
Qwen3.8-27B & \texttt{Qwen/Qwen3.8-27B} & Alibaba & 27B$^{d}$ & served endpoint & unreported & \texttt{1d4bf0f2}$^{c}$ \\
GLM-5.2 & \texttt{zai-org/GLM-5.2-FP8} & Z.ai & 753B MoE & served endpoint & FP8$^{e}$ & \texttt{f33c6dc5}$^{c}$ \\
\bottomrule
\end{tabular}
}
\caption{\textbf{One panel of seven models from six families.} We use the short names in the first column throughout. Revisions are Hugging Face commits. $^{a}$Includes the vision wrapper; only text inputs are used. $^{b}$Per the DeepSeek-V4 report; the Hugging Face listing for the 0731 checkpoint gives 304B because it includes an attached speculative-decoding module. $^{c}$Revision of the tokenizer and chat template we pin; the endpoint does not expose its weight revision. $^{d}$Dense; 28B with its vision encoder. $^{e}$Named by the served model alias and not independently verified. MoE: mixture of experts; we do not report active parameters.}
\label{tab:models}
\end{table*}

Table~\ref{tab:models} lists the checkpoints. Every model generates and evaluates its own outputs. The four local models run in bfloat16 without quantization on one NVIDIA A10 each, with Python 3.12, PyTorch 2.6.0, CUDA 12.6, and Transformers 4.57.6. The 30-item Llama-3.1-8B ladder and headroom runs use Transformers 4.48.3. The three served models are queried with at most four concurrent requests.

In total, we used approximately 15 NVIDIA A10 GPU-hours for the local-model experiments and 16 million input and output tokens across the three served models. These estimates cover generation and likelihood scoring for the main experiments, excluding pilot experiments and reruns.

\subsection{Scoring conventions}
\label{sec:scoring}

Each model uses its native chat template with the system message \texttt{You are a helpful assistant.}, and optional thinking modes are disabled. Conditional scoring concatenates the formatted prefix, the response, and EOS, masks every prefix label, and sums the negative log-likelihood over the response-plus-EOS positions. Unconditional scoring evaluates the same response-plus-EOS tokens after one boundary token, which is BOS where the tokenizer defines one. Qwen3-8B, Qwen3.8-27B, and GLM-5.2 define no BOS and use EOS as the left boundary. Because a fixed output has the same unconditional denominator under every context, same-output contrasts do not depend on this convention. Absolute scores remain model-relative and are not calibrated across families. Scores are never clipped.

On served endpoints we obtain teacher-forced prompt log-probabilities and check every returned token ID against the pinned tokenizer. The one completion token the endpoint must generate is excluded. For GLM-5.2 we disable automatic special-token insertion because the native template already supplies its control tokens. For fixed suffixes, character offsets are mapped to token offsets so that body and suffix log-likelihoods can be separated. The output-only estimate uses the released Llama-3.1-8B threshold $-\log\tau=0.5496$. No thresholds were released for other models, so we report the estimate for Llama-3.1-8B only.

\subsection{Decoding, seeds, and statistics}
\label{sec:stats}

Generation uses temperature 0.7, top-$p$ 0.9, and at most 384 new tokens, with one clean and one attacked generation per item. Served runs seed generation with $17+i$ for clean and $10017+i$ for attacked outputs at manifest index $i$. Local runs use a seeded batched sampling stream, and run metadata record every seed. The 384-token limit is reached by 16/300 clean and 25/300 attacked DeepSeek-V4-Flash news generations, 50/300 and 148/300 for Qwen3.8-27B, and 2/300 and 0/300 for GLM-5.2 (arXiv counts are in Appendix~\ref{sec:arxiv}). All outputs are retained, including truncated ones. Every interval is a 95\% percentile interval from 10,000 bootstrap resamples of items, with all paired conditions kept together, and no example is discarded after generation.

\section{Additional Results}
\label{sec:more-results}

This appendix shows the estimates behind each claim of Section~\ref{sec:results}. We give per-model intervals, the dose response for every run shown, the Llama-3.1-8B diagnostics, the arXiv replication, and three subtler lexical variants that the ledger predicts should barely work (they barely do).

\subsection{Per-model effects}
\label{sec:per-model}

\begin{table*}[t]
\centering
\scriptsize
\resizebox{\textwidth}{!}{\begin{tabular}{@{}lccccc@{}}
\toprule
& \multicolumn{3}{c}{Fixed output} & \multicolumn{2}{c}{Generation-time} \\
\cmidrule(lr){2-4}\cmidrule(l){5-6}
Model & \ctx{}$-$clean & \ctx{}$-$\hid{} & \ctx{}$-$\wrg{} & Attack$-$clean & \ctx{}$-$\hid{} \\
\midrule
Llama-3.1-8B & +0.130 [+0.125, +0.135] & +0.505 [+0.500, +0.510] & +0.252 [+0.250, +0.254] & +0.199 [+0.185, +0.213] & +0.521 [+0.490, +0.551] \\
Qwen3-8B & -0.144 [-0.150, -0.138] & +0.333 [+0.327, +0.339] & +0.293 [+0.290, +0.297] & +0.146 [+0.142, +0.150] & +0.793 [+0.783, +0.802] \\
Gemma-3-4B & -0.062 [-0.073, -0.052] & +0.339 [+0.330, +0.347] & +0.231 [+0.228, +0.234] & +0.235 [+0.224, +0.245] & +0.843 [+0.831, +0.855] \\
Phi-4-mini & +0.211 [+0.206, +0.215] & +0.388 [+0.383, +0.392] & +0.224 [+0.221, +0.226] & +0.271 [+0.261, +0.282] & +0.505 [+0.494, +0.516] \\
DeepSeek-V4-Flash & -0.047 [-0.051, -0.042] & +0.408 [+0.402, +0.415] & +0.241 [+0.238, +0.244] & +0.133 [+0.128, +0.139] & +0.701 [+0.690, +0.711] \\
Qwen3.8-27B & +0.095 [+0.091, +0.099] & +0.324 [+0.318, +0.331] & +0.163 [+0.160, +0.167] & +0.161 [+0.150, +0.171] & +0.506 [+0.491, +0.521] \\
GLM-5.2 & +0.072 [+0.065, +0.080] & +0.426 [+0.420, +0.432] & +0.248 [+0.245, +0.252] & +0.215 [+0.206, +0.224] & +0.621 [+0.613, +0.629] \\
\bottomrule
\end{tabular}
}
\caption{\textbf{Every model moves, and every interval excludes zero.} Paired mean effects at eight IDs with 95\% bootstrap intervals, 300 news items per model. Fixed-output contrasts score one padded text under two contexts, except \ctx{}$-$clean, which also changes the text. Attack$-$clean compares separately generated articles. Generation-time \ctx{}$-$\hid{} rescores the same attacked article with the full prompt and with the visible request.}
\label{tab:multimodel-effects}
\end{table*}

No single model drives the averages (Table~\ref{tab:multimodel-effects}). Showing the matching IDs raises the fixed-output score by \CrossFixedMin{}--\CrossFixedMax{} and does so in all 2,100 pairs, and \ctx{} beats \wrg{} in every pair as well. Generation-time attack$-$clean effects are positive for 2,040 of 2,100 articles, and generation-time \ctx{}$-$\hid{} for 2,097 of 2,100. Exact ID recall in attacked articles is 0.65 (Llama-3.1-8B), 0.98 (Qwen3-8B), 0.98 (Gemma-3-4B), 0.85 (Phi-4-mini), 0.94 (DeepSeek-V4-Flash), 0.93 (Qwen3.8-27B), and 0.99 (GLM-5.2). Mean clean/attacked word counts, in the same order, are 216/198, 232/195, 220/204, 263/273, 229/222, 281/265, and 227/209. The generation-time effect therefore does not track length, and the fixed-output contrasts remove length altogether.

\subsection{Dose response}
\label{sec:dose}

\begin{figure*}[t]
\centering
\includegraphics[width=0.61\textwidth]{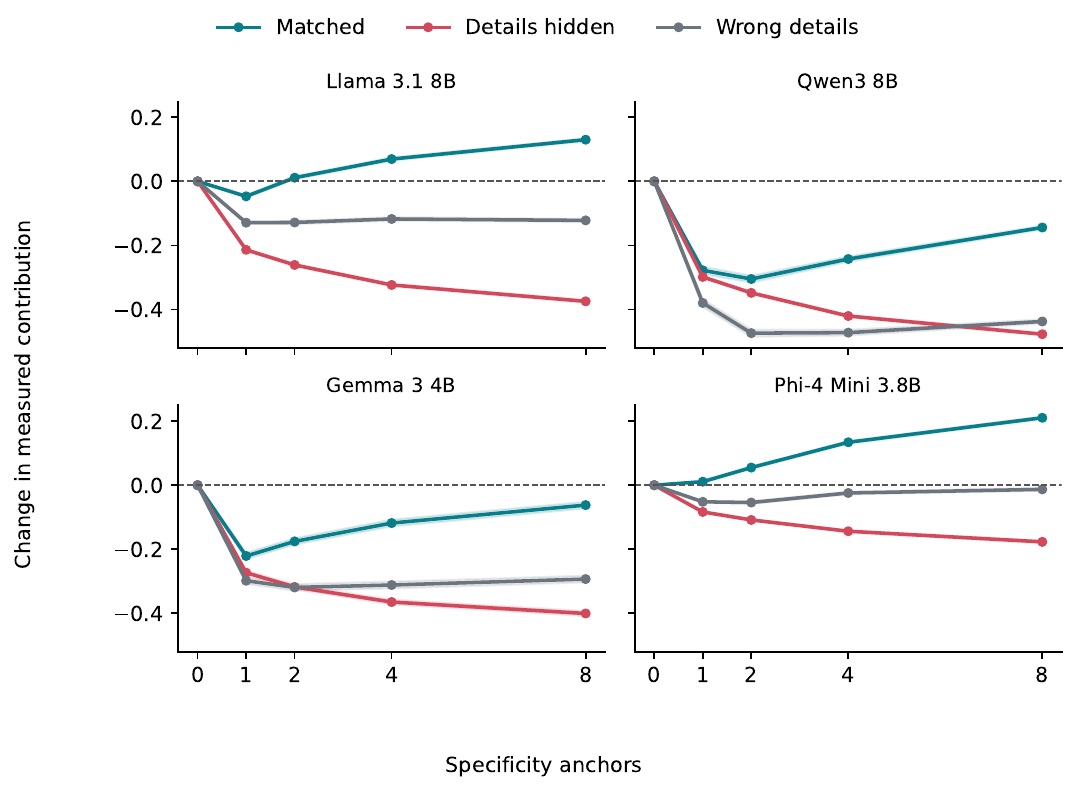}\hfill
\begin{minipage}[b]{0.37\textwidth}
\centering
\includegraphics[width=\linewidth,trim=0 0 0 21,clip]{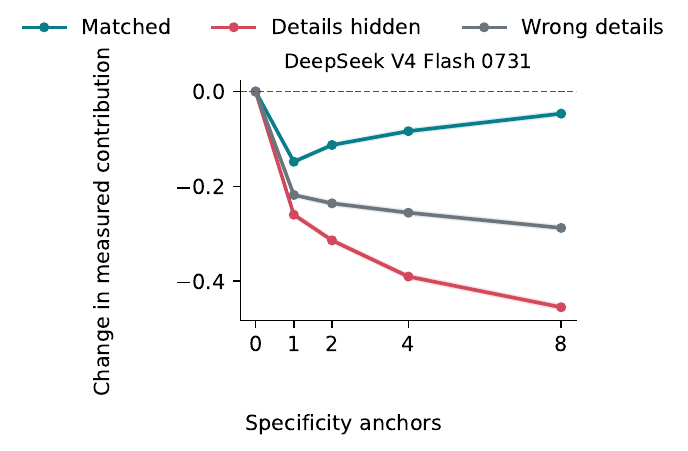}\\[2pt]
\includegraphics[width=\linewidth,trim=0 0 0 21,clip]{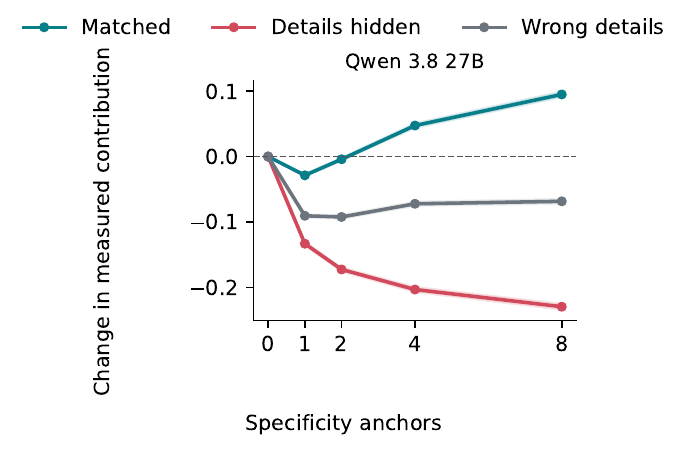}
\end{minipage}\\[4pt]
\includegraphics[width=0.86\textwidth,trim=0 0 0 21,clip]{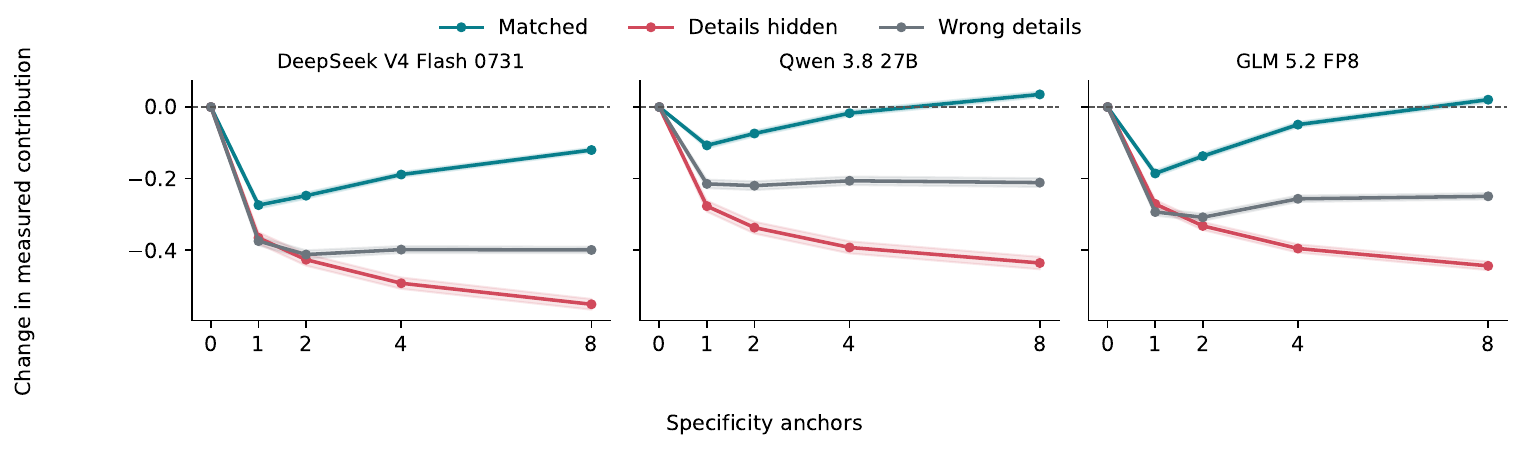}
\caption{\textbf{\ctx{} stays above \hid{} and \wrg{} at every dose, for every run shown.} Fixed-output change from the clean score against the number of IDs $K$ ($K=0$ is the clean output). At each positive dose the three contexts score the same padded text. Top: news, 300 items per model. The GLM-5.2 news dose curve is not shown, and its eight-ID estimates are in Table~\ref{tab:multimodel-effects}. Bottom: arXiv, 200 items per model. Bands are paired 95\% bootstrap intervals. The embedded labels predate our naming scheme. ``Details hidden'' is \hid{}, ``Wrong details'' is \wrg{}, ``Specificity anchors'' counts IDs, and panel titles abbreviate the names in Table~\ref{tab:models}.}
\label{fig:dose}
\end{figure*}

Corollary~\ref{cor:padding} predicts the shape of these curves. As IDs accumulate, the suffix's surprisal share $w$ grows, so \ctx{} scores move toward the suffix's local score and \hid{} scores fall. Figure~\ref{fig:dose} shows both trends in every run. Where \ctx{} first dips below clean at one ID (every run shown except Phi-4-mini), the pattern is consistent with the first IDs perturbing the body while the short suffix is still dominated by its unpredictable metadata header. Added IDs then outweigh both, and the curves climb.

\subsection{Llama-3.1-8B diagnostics}
\label{sec:llama-diag}

\begin{figure*}[t]
\centering
\includegraphics[width=0.55\textwidth]{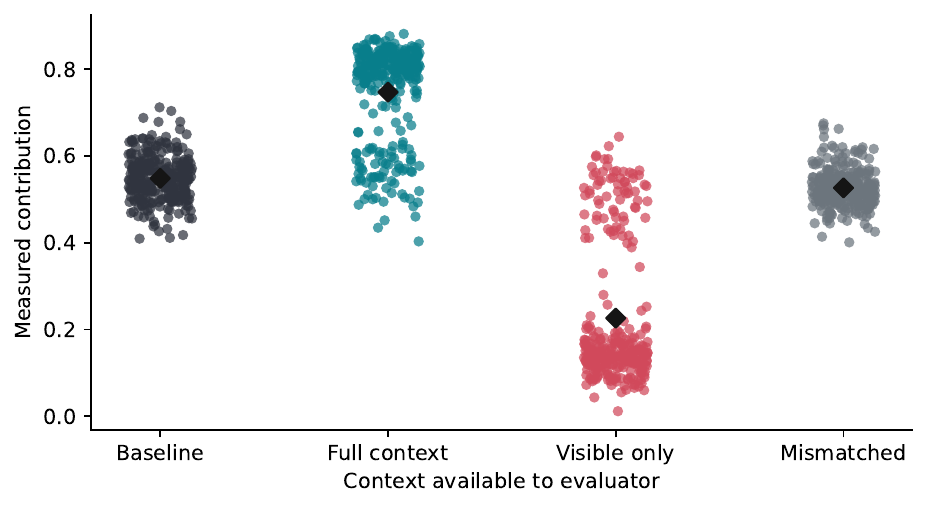}\hfill
\includegraphics[width=0.40\textwidth,trim=262 0 0 56,clip]{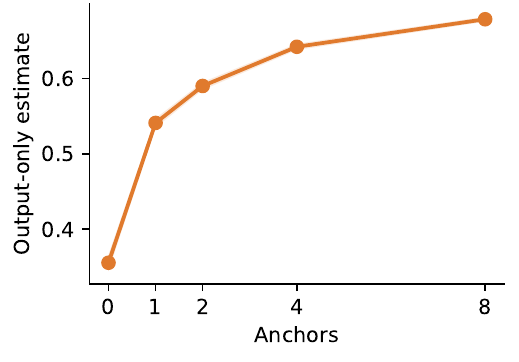}
\caption{\textbf{Two ways to inflate one model's score.} Llama-3.1-8B, all 300 news items. Left: generation-time articles scored under each evaluator context. The same attacked articles average 0.75 under \ctx{} and 0.23 under \hid{}; diamonds are means. Axis labels ``Baseline'', ``Full context'', ``Visible only'', and ``Mismatched'' denote the clean article, \ctx{}, \hid{}, and \wrg{}. Right: the output-only estimate of fixed padded outputs rises with the number of IDs (axis ``Anchors'') although no prompt is used. Bands are paired 95\% bootstrap intervals.}
\label{fig:llama-diag}
\end{figure*}

\begin{table}[t]
\centering
\small
\resizebox{\columnwidth}{!}{\begin{tabular}{lrrrr}
\toprule
Evaluator context & Mean score & $\Delta$ & 95\% CI & Recall \\
\midrule
\ctx{} (full prompt) & 0.747 & +0.199 & [+0.185, +0.213] & 0.65 \\
\hid{} (visible request) & 0.226 & -0.322 & [-0.340, -0.304] & 0.65 \\
\wrg{} (wrong IDs) & 0.526 & -0.022 & [-0.028, -0.016] & 0.65 \\
\bottomrule
\end{tabular}
}
\caption{\textbf{Hiding the IDs erases the attack and more.} Llama-3.1-8B generation-time articles with eight requested IDs, 300 items. $\Delta$ is relative to the separately generated clean article, and recall is exact ID recall in the output.}
\label{tab:natural}
\end{table}

\begin{table}[t]
\centering
\small
\resizebox{\columnwidth}{!}{\begin{tabular}{lrrrr}
\toprule
Assistance level & Clean score & \ctx{} $\Delta$ & \hid{} $\Delta$ & \wrg{} $\Delta$ \\
\midrule
Polish (full draft) & 0.846 & -0.004 & -0.364 & -0.172 \\
Summary (key ideas) & 0.708 & +0.054 & -0.397 & -0.159 \\
Title & 0.536 & +0.137 & -0.366 & -0.112 \\
Subject & 0.488 & +0.143 & -0.353 & -0.120 \\
\bottomrule
\end{tabular}
}
\caption{\textbf{\ctx{} inflation vanishes as the clean score nears the suffix's local score.} Eight-ID fixed-output shifts from the clean score across assistance levels (Llama-3.1-8B, 30-item subset; Figure~\ref{fig:headroom}).}
\label{tab:cross-mode}
\end{table}

Llama-3.1-8B is the one model with a released output-only threshold and the model behind the ladder, so we report two further views of it. In generation, attacked articles score 0.747 under \ctx{}. Removing only the IDs from the context drops the same articles to 0.226, a paired gap of 0.521 (95\% CI [0.490, 0.551]), and \wrg{} yields 0.526 (Table~\ref{tab:natural}, Figure~\ref{fig:llama-diag}, left). With no prompt at all, the output-only estimate rises from 0.36 for clean outputs to 0.68 at eight IDs (Figure~\ref{fig:llama-diag}, right). Table~\ref{tab:cross-mode} gives the fixed-output shifts behind Figure~\ref{fig:headroom}.

\subsection{Replication on arXiv abstracts}
\label{sec:arxiv}

\begin{table*}[t]
\centering
\scriptsize
\begin{tabular}{lccc}
\toprule
Model & \ctx{}$-$clean & \ctx{}$-$\hid{} & \ctx{}$-$\wrg{} \\
\midrule
DeepSeek-V4-Flash & $-0.120$ $[-0.127,-0.114]$ & $+0.431$ $[+0.417,+0.445]$ & $+0.279$ $[+0.274,+0.284]$ \\
Qwen3.8-27B & $+0.035$ $[+0.027,+0.043]$ & $+0.471$ $[+0.459,+0.483]$ & $+0.246$ $[+0.241,+0.252]$ \\
GLM-5.2 & $+0.021$ $[+0.013,+0.028]$ & $+0.465$ $[+0.454,+0.475]$ & $+0.270$ $[+0.265,+0.275]$ \\
\bottomrule
\end{tabular}

\vspace{6pt}
\begin{tabular}{lcccc}
\toprule
Model & Attack$-$clean & \ctx{}$-$\hid{} & \ctx{}$-$\wrg{} & Recall \\
\midrule
DeepSeek-V4-Flash & $+0.134$ $[+0.128,+0.140]$ & $+0.778$ $[+0.769,+0.788]$ & $+0.364$ $[+0.357,+0.370]$ & 0.996 \\
Qwen3.8-27B & $+0.196$ $[+0.188,+0.204]$ & $+0.663$ $[+0.649,+0.676]$ & $+0.287$ $[+0.279,+0.296]$ & 0.991 \\
GLM-5.2 & $+0.196$ $[+0.188,+0.204]$ & $+0.678$ $[+0.669,+0.686]$ & $+0.300$ $[+0.295,+0.306]$ & 0.995 \\
\bottomrule
\end{tabular}

\caption{\textbf{The audit replicates on scientific abstracts.} Eight IDs, $n=200$ per model, paired mean differences with 95\% bootstrap intervals. Top, fixed output: \ctx{}$-$\hid{} and \ctx{}$-$\wrg{} compare identical padded outputs, while \ctx{}$-$clean also changes the output. Bottom, generation time: attack$-$clean compares separately generated outputs, the other two contrasts rescore the same attacked output, and recall is exact ID recall.}
\label{tab:arxiv}
\end{table*}

The arXiv runs repeat both interventions with the three served models (600 model--abstract pairs), using the settings of Appendix~\ref{sec:details}. Showing the matching IDs raises fixed-output scores by 0.431--0.471, and \ctx{} beats \wrg{} by 0.246--0.279, in all 600 pairs (Table~\ref{tab:arxiv}). \ctx{} padding stays below the clean reference for DeepSeek-V4-Flash ($-0.120$), whose clean abstracts already score 0.79, and exceeds it for Qwen3.8-27B ($+0.035$) and GLM-5.2 ($+0.021$), as the headroom account of \S\ref{sec:mechanism} predicts. Generation-time attack$-$clean effects are $+0.134$, $+0.196$, and $+0.196$, positive for 200, 199, and 200 of 200 abstracts. Generation-time \ctx{}$-$\hid{} effects range from 0.663 to 0.778 and \ctx{}$-$\wrg{} effects from 0.287 to 0.364, positive in all 600 pairs. ID recall exceeds 0.99. Mean clean/attacked word counts are 133.66/124.74 (DeepSeek-V4-Flash), 161.37/153.00 (Qwen3.8-27B), and 135.75/118.71 (GLM-5.2). The 384-token limit is reached by 6/6, 18/31, and 3/3 of 200 clean/attacked abstracts, and all are retained. Differences from the news effects are descriptive, since items and output lengths differ.

\subsection{Subtler lexical variants}
\label{sec:lexical}

\begin{table}[!htb]
\centering
\small
\resizebox{\columnwidth}{!}{%
\begin{tabular}{@{}lc@{}}
\toprule
Variant (vs.\ clean, points) & Mean [95\% CI] \\
\midrule
\multicolumn{2}{@{}l}{\emph{Two required words in the prompt}} \\
\quad Common words & $+3.0$ [$+1.9$, $+4.1$] \\
\quad Random rare words & $+3.2$ [$+2.2$, $+4.2$] \\
\quad Score-selected rare words & $+3.5$ [$+2.5$, $+4.4$] \\
\multicolumn{2}{@{}l}{\emph{One required three-token phrase}} \\
\quad Surprisal-selected & $+3.4$ [$+2.4$, $+4.4$] \\
\quad Surprisal-selected $-$ random phrase & $+0.6$ [$-0.4$, $+1.6$] \\
\multicolumn{2}{@{}l}{\emph{Up to two rarer synonyms near the end, prompt fixed}} \\
\quad Random & $-5.1$ [$-5.8$, $-4.4$] \\
\quad Rarity-greedy & $-6.2$ [$-7.0$, $-5.5$] \\
\quad Score-greedy & $-0.2$ [$-0.4$, $-0.0$] \\
\bottomrule
\end{tabular}}
\caption{\textbf{A few rare words buy almost nothing.} Lexical pilots with DeepSeek-V4-Flash ($n=60$ news items, paired bootstrap intervals). Score-greedy editing keeps the original text when no candidate helps, which happened for 47 items.}
\label{tab:lexical}
\end{table}

We also piloted three less conspicuous manipulations with DeepSeek-V4-Flash on 60 predetermined news items (Table~\ref{tab:lexical}). Two required words in the prompt, or one required three-token phrase, raise the score by about 3 points over the clean prompt, and rarity-based selection does not beat random or common choices. Replacing up to two words near the end of a fixed article with rarer synonyms, with the prompt unchanged, lowers the score. Proposition~\ref{prop:ledger} predicts both outcomes. A few copyable words carry little surprisal, so they earn little credit. Rare words that the prompt does not predict have local scores near zero, so they pull the ledger down. The same rare edits raise the output-only estimate in 58 and 59 of 60 articles for the random and rarity-greedy strategies, as Corollary~\ref{cor:output-only} predicts. Repeated scoring of identical texts on this endpoint had a mean within-text standard deviation of 0.65 points (maximum 1.29), so differences of a point or less are not interpretable.

\section{Theory}
\label{sec:theory}

This appendix proves the results of Sections~\ref{sec:mechanism} and~\ref{sec:identification}, then puts the ledger's quantitative predictions to the Llama-3.1-8B estimates. The checks use aggregate scores only, so they bound the suffix's surprisal share rather than measure it span by span.

\subsection{Proofs}
\label{sec:proofs}

Throughout, the scored sequence is the response followed by EOS, and the terminal token belongs to the last span.

\paragraph{Proposition~\ref{prop:ledger}.} By the chain rule, $\surp(y)=\sum_k S_k$ and $\surp(y\mid x)=\sum_k T_k$. Hence
\begin{align*}
\score(x,y)&=\frac{\sum_k S_k-\sum_k T_k}{\surp(y)}\\
&=\sum_k\frac{S_k-T_k}{\surp(y)}=\sum_k\kappa_k.
\end{align*}
Under a softmax model every nonempty span has $S_k>0$, so $\kappa_k=(S_k/\surp(y))(1-T_k/S_k)=w_k\score_k$, and $\sum_k w_k=1$. For the output-only estimate, $\widehat{\score}(y)=1-bN/\surp(y)=\sum_k (S_k-b\,n_k)/\surp(y)$ because $\sum_k n_k=N$.

\paragraph{Corollary~\ref{cor:padding}.} Apply Proposition~\ref{prop:ledger} to the two spans $y$ and $z$ under $x'$. The first span's local score is $1-\surp(y\mid x')/\surp(y)=\score(x',y)$ and its weight is $1-w$. The second span has weight $w$ and local score $\score_z(x')$, which gives Equation~\ref{eq:padding}. Since $0<w<1$, the padded score lies between the two local scores, and $\score(x',yz)-\score(x',y)=w\,[\score_z(x')-\score(x',y)]$ is positive exactly when $\score_z(x')>\score(x',y)$. Finally,
\begin{equation*}
\score(x',yz)=1-\frac{\surp(y\mid x')+\surp(z\mid x',y)}{\surp(y)+\surp(z\mid y)},
\end{equation*}
and the fraction tends to zero when $\surp(z\mid y)\to\infty$ with $\surp(y)$ fixed and the numerator $o(\surp(z\mid y))$.

\paragraph{Context swaps.} For a fixed output $\tilde{y}$ and contexts $x_1,x_0$, the denominators of Equation~\ref{eq:score} coincide, so $\score(x_1,\tilde{y})-\score(x_0,\tilde{y})=[\surp(\tilde{y}\mid x_0)-\surp(\tilde{y}\mid x_1)]/\surp(\tilde{y})$. Applying Equation~\ref{eq:padding} under both contexts and subtracting gives Equation~\ref{eq:gap}. Both terms share the weight $w$ because unconditional surprisals do not depend on the context.

\paragraph{Corollary~\ref{cor:output-only}.} With $b>0$, $\frac{d}{d\bar{\surp}}\big(1-b/\bar{\surp}\big)=b/\bar{\surp}^{2}>0$, and $1-b/\bar{\surp}\to1$ as $\bar{\surp}\to\infty$. If an $n$-token output has mean surprisal $\bar L$ and an appended $m$-token span has mean surprisal $\bar S>\bar L$ given the output, the new mean $(n\bar L+m\bar S)/(n+m)$ exceeds $\bar L$.

\paragraph{Proposition~\ref{prop:nonident}.} Let $c_-=\inf\mathcal{C}(o)$, $c_+=\sup\mathcal{C}(o)$, and $a=f(o)$. For any $\epsilon>0$, choose compatible values within $\epsilon$ of $c_-$ and $c_+$. By the triangle inequality, the two absolute errors sum to at least $c_+-c_--2\epsilon$, so the larger is at least $D(o)/2-\epsilon$. Let $\epsilon\downarrow0$. The midpoint $a^*=(c_-+c_+)/2$ satisfies $|a^*-c|\leq D(o)/2$ for every $c\in\mathcal{C}(o)$. For a randomized estimator $A$, choose $c_1,c_2\in\mathcal{C}(o)$ with $c_2-c_1\geq D(o)-2\epsilon$. Then $\mathbb{E}|A-c_1|+\mathbb{E}|A-c_2|\geq c_2-c_1$, so one expected error is at least $D(o)/2-\epsilon$. For the laundering statement, both $0$ and $c_1$ lie in $\mathcal{C}(o)$, so $D(o)\geq c_1$.

\paragraph{Scope.} The target $c$, the observation map $g$, and the admissible history class jointly determine $\mathcal{C}(o)$. Authenticated logs or workflow constraints shrink this set, and selectively exported transcripts enlarge it. The result permits prior-based prediction, population estimation, and distributional detection. It states only that transcript-level point identification requires $c$ to be constant on each observational-equivalence class.

\subsection{Mechanism checks}
\label{sec:mechanism-appendix}

\paragraph{How much surprisal do eight IDs carry?} By Equation~\ref{eq:gap}, the \ctx{}$-$\hid{} gap approximates the suffix weight $w$ when showing the IDs makes them near-certain ($\score_z^{\mathrm{M}}$ near one), hiding them leaves them unpredictable ($\score_z^{\mathrm{H}}$ near zero), and the body term is small. Llama-3.1-8B's output-only estimates give an independent handle on $w$. Inverting Equation~\ref{eq:output-only} with $b=0.5496$, the mean estimates for clean outputs (0.356) and eight-ID padded outputs (0.679) correspond to mean surprisals of about 0.85 and 1.71 nats per token (harmonic means across items). Appending $m$ suffix tokens to an $N$-token body gives
\begin{equation*}
w=1-\frac{N}{N+m}\cdot\frac{\bar{\surp}(y)}{\bar{\surp}(yz)}\;\geq\;1-\frac{0.85}{1.71}\approx0.50,
\end{equation*}
ignoring item heterogeneity and the terminal token. Both routes indicate that the eight IDs carry about half of the padded document's surprisal, although they occupy a minority of its tokens.

\paragraph{Headroom.} Regressing the \ctx{} shift in Table~\ref{tab:cross-mode} on the clean score gives slope $-0.43$, intercept $0.36$, a zero crossing at $0.84$, and $R^2=0.99$ over the four levels. Under Corollary~\ref{cor:padding}, if the IDs leave the body's own score unchanged, the \ctx{} shift is $w(\score_z^{\mathrm{M}}-\score_0)$, so inflation vanishes when the clean score $\score_0$ reaches the suffix's local score. A change $\delta$ in the body's score adds $(1-w)\delta$ and moves the crossing accordingly. Because the \hid{} context equals the clean context, the \hid{} shift is exactly $w(\score_z^{\mathrm{H}}-\score_0)$, up to the terminal token. Its near-constancy across levels ($-35$ to $-40$ points) implies that $w$ shrinks as the clean score rises, as expected when polished outputs follow a human-written article and therefore carry more surprisal in the body. Span-level measurements would separate $w$ from the local scores.

\end{document}